\pdfoutput=1
\documentclass{tADR2e}

\usepackage{bm}
\usepackage[caption=false]{subfig}
\usepackage{algpseudocodex}
\usepackage[hidelinks]{hyperref}

\theoremstyle{plain}

\theoremstyle{definition}

\theoremstyle{remark}

\newcommand{\secref}[1]{Section~\ref{#1}}
\newcommand{\figref}[1]{Figure~\ref{#1}}

\title{\LARGE \textbf{Design optimization of tendon-driven robots considering tendon wrapping and shortcut}}

\author{Yuta Sahara$^{a}$, Kento Kawaharazuka$^{a,b}$$^{\ast}$, and Kei Okada$^{a}$%
  \thanks{$^\ast$Corresponding author. Email: kawaharazuka@jsk.imi.i.u-tokyo.ac.jp \vspace{6pt}}\\\vspace{6pt}
  $^{a}${\em{Department of Mechano-Informatics, Graduate School of Information Science and Technology, The University of Tokyo, 7-3-1 Hongo, Bunkyo-ku, Tokyo, 113-8656, Japan.}}\\\vspace{6pt}
  $^{b}${\em{AI Center, Graduate School of Information Science and Technology, The University of Tokyo, Japan.}}
}

\begin{document}

\makeatletter
\let\ps@plain\ps@empty
\def\@received{Published online: 31 August 2026}
\makeatother

\maketitle

\begin{center}
  \small
  This is the author manuscript of an article published in \emph{Advanced Robotics} (2026),
  published by Taylor \& Francis on behalf of the Robotics Society of Japan.
  The final published version is available at
  \url{https://doi.org/10.1080/01691864.2026.2723650}.
\end{center}

\begin{abstract}
	This study proposes a multi-objective optimization method for tendon-driven arm design, addressing the inherent trade-off between joint torque and arm thickness through tendon wrapping and shortcut effects. We formulate the problem to simultaneously maximize torque and minimize thickness, solved using NSGA-II. The algorithm optimizes tendon routing, pulley configurations, and attachment points. Results reveal Pareto-optimal designs with effective moment arm expansion via strategic shortcuts, providing practical guidelines for compact high-torque arms. The optimized configurations demonstrate non-trivial patterns beyond conventional design intuition, offering new insights for engineering efficient robotic systems.
\end{abstract}

\begin{keywords}
	tendon-driven robot; design optimization; multi-objective optimization
\end{keywords}

\section{Introduction}
  
  Thus far, various tendon-driven robots have been developed. 
  Among these, musculoskeletal humanoid robots such as ECCE1\cite{ECCE1}, Anthrob\cite{Anthrob}, Kengoro\cite{Kengoro}, and Musashi\cite{Musashi} represent bio-inspired designs, 
  while non-biomimetic wire-driven systems like Super Dragon\cite{SuperDragon}, SAQIEL\cite{SAQIEL}, and Twist Snake\cite{TwistSnake} have also been investigated through research and development efforts.
	Among these diverse forms, this study focuses on serial-link tendon-driven arms with pulleys, where high joint torque is required under space-limited conditions.
	This class includes both musculoskeletal and non-biomimetic wire-driven systems, and in such systems the tendon-routing geometry strongly influences torque capability and arm thickness.
	Therefore, our scope is defined by routing geometry rather than by a specific actuation source.
  
  Tendon-driven systems offer several advantages. 
  First, increasing the moment arm enables torque amplification. 
  Additionally, separating actuation units from joints enhances design flexibility in robotic configurations. 
  Furthermore, control benefits include variable stiffness control \cite{kobayashi1998tendon} and redundancy against tendon failure \cite{KAWAHARAZUKA2022104067}. 
  However, maximizing these advantages requires careful consideration of tendon placement. 
  Designing robots that fully exploit tendon-driven mechanisms is therefore non-trivial. 
  Specifically, elements such as tendon pathways and pulley design significantly influence overall robotic performance. 
  Consequently, designers must meticulously evaluate these factors to achieve optimal system configurations.
  
  \begin{figure}[t]
	  \centering
	  \includegraphics[width=0.5\columnwidth]{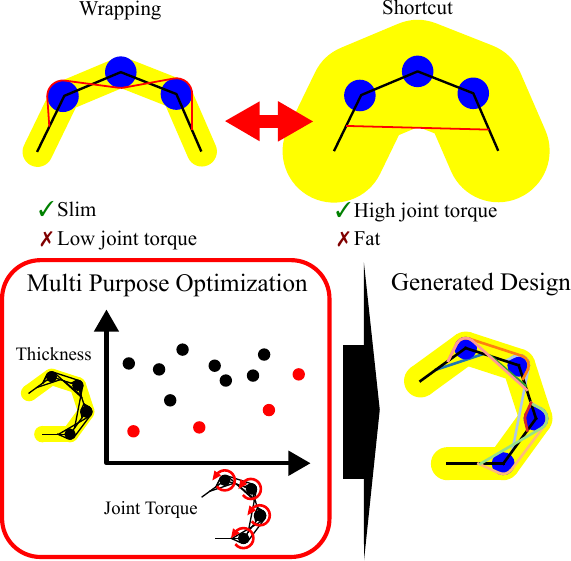}
	  \caption{The overview of this study.
		  Design optimization of tendon-driven arms that take advantage of tendon wrapping and shortcuts through multi-objective optimization that considers trade-offs in joint torque and arm thickness.}
	  \label{fig:abst}
  \end{figure}
  
  Based on these considerations, studies have explored methodologies for optimizing tendon configurations.   
  Tsai et al. \cite{Tsai1996Isotropic} proposed an isotropic design method for tendon-driven manipulators 
  to ensure uniform force transmission characteristics, providing fundamental design guidelines for tendon routing.
  In \cite{2012Asaoka}, for the leg section of a musculoskeletal robot, the muscle pathways per pulley were analyzed and optimized; 
  however, this study fixed the moment arm and did not account for changes in moment arms due to joint angle variations.   
  \cite{2024Kawaharazuka} further optimized tendon configurations by determining optimal positions for start points, endpoints, and intermediate via-points of tendons during operation, explicitly considering dynamic changes in moment arms caused by joint movements.   
  However, among methods to amplify moment arms more effectively, designs that leverage tendon wrapping around pulleys and tendon shortcuts are notable. 
  In such designs, moment arms are maintained through tendon-wrapping on pulleys. Additionally, tendon shortcuts—caused by angular changes in joints—further expand the effective moment arm beyond the pulley's radius, thereby enhancing joint torque.   
  Despite their potential, optimization studies incorporating these phenomena remain limited in prior research. 
  This is likely due to the computational complexity involved: 
  calculating tendon pathways for each joint angle while accounting for wrapping and shortcut transitions requires intricate modeling.   
  Furthermore, practical challenges arise in robotic arm applications. 
  Tendon shortcuts can increase arm thickness, restricting achievable poses and operational workspace. 
  For robots like \cite{SuperDragon}, which operate in extremely confined spaces, minimizing arm diameter is critical to ensure maneuverability.   
  Thus, designs exploiting tendon wrapping/shortcuts face a trade-off between increasing joint torque and reducing arm thickness. 
  Engineers must navigate this dilemma: maximizing torque through dynamic moment arms inherently conflicts with the need for slender arms in narrow operational environments. 
  This necessitates innovative solutions to simultaneously optimize torque enhancement and compact design—a central challenge in tendon-driven robotic systems. 
  
  Based on these considerations, this study proposes a design optimization method incorporating tendon shortcuts for serial-link tendon-driven arms with pulleys. 
  Specifically, we address two objectives:
  \begin{enumerate}
	  \item Enhancing joint torque by leveraging moment arm enlargement caused by tendon shortcuts in the tendon pathways, and
	  \item Minimizing arm thickness by preventing excessive tendon shortcuts.
  \end{enumerate}
  To achieve these goals, this research conducts a multi-objective optimization of tendon configurations and routing designs,
  explicitly modeling both torque enhancement via shortcut-induced moment arms and geometric constraints to avoid unnecessary arm thickening. 
  The effectiveness of the proposed method is demonstrated through evaluating optimized design candidates generated under these dual objectives. 
  
  The following sections outline the structure of this study.
  \secref{sec:method} describes the design parameters of the tendon-driven arm to be optimized, 
  explains computational methods for tendon pathways and muscle length Jacobian, 
  details evaluation methodologies for joint torque and arm thickness, and outlines the multi-objective optimization framework employed.   
  In \secref{sec:experiments}, we conduct multi-objective optimizations by varying the number of joints and tendons in the system, presenting the results of these optimizations.
  \secref{sec:discussion} critically examines the optimization outcomes to demonstrate the effectiveness of our proposed method through systematic analysis. 
  Finally, concluding remarks summarizing the study's contributions are provided in \secref{sec:conclusion}. 

\section{METHOD}
  
  \label{sec:method}
  \subsection{Overview of Optimization}
	  The problem formulation of this study is described as follows.
	  In this research, we address an optimization problem concerning tendon configurations in a two-dimensional serial-link arm driven by multiple tendons. 
	  Specifically, maximizing the distance between each tendon and joints—the moment arm—enables enhanced capability to generate joint torque. 
	  Achieving this objective necessitates careful design of tendon configurations. 
	  For instance, increasing the moment arm can be achieved not by wrapping all tendons around every pulley but rather through strategically shortcutting certain pulleys. 
	  However, this short-cutting of tendons also results in an increase in the overall thickness of the arm. 
	  This creates a trade-off between improving joint torque and minimizing arm thickness.
	  This study formulates an optimization problem for tendon configurations while explicitly considering this trade-off relationship.
  \subsection{The design parameters of the arm model}
	  \label{subsec:arm_design}
	  \newcommand{\njoint}{N_{\text{joint}}}
	  \newcommand{\ntendon}{N_{\text{tendon}}}
	  \newcommand{\link}[1]{{link}_{#1}}
	  \newcommand{\joint}[1]{{joint}_{#1}}
	  \newcommand{\qpos}[1]{\theta_{#1}}
	  \newcommand{\qposList}{\bm{\theta}}
	  \newcommand{\tendon}[1]{{tendon}_{#1}}
	  \newcommand{\tendonStart}[1]{S^\text{start}_{#1}}
	  \newcommand{\tendonEnd}[1]{S^\text{end}_{#1}}
	  \newcommand{\tendonStartX}[1]{P^\text{start}_{#1}}
	  \newcommand{\tendonEndX}[1]{P^\text{end}_{#1}}
	  \newcommand{\tendonSide}[2]{{tendon\_side}_{#1,#2}}
	  \newcommand{\NaturalNumber}{\mathbb{N}}
	  \newcommand{\RealNumber}{\mathbb{R}}
	  In this section, we describe the design parameters and model configuration of the arm subject to this study. 
	  The number of joints is denoted as $\njoint\in\NaturalNumber$, and the number of tendons as $\ntendon\in\NaturalNumber$. 
	  In \figref{fig:model}, a specific example with $\njoint=3$ joints and $\ntendon=5$ tendons is illustrated. 
	  Arm links are sequentially named from proximal to distal as $\link{0}, \link{1},\dots, \link{\njoint}$. 
	  The length of each link is represented by $L_i\in\RealNumber$. 
	  Joints are designated as $\joint{0}, \joint{1},\dots, \joint{\njoint-1}$ with their angles denoted by $\qpos{i}$. 
	  We define $\qpos{i}=0$ when the arm is in-line, adopting counterclockwise direction as the positive orientation. 
	  Joint angles are collectively expressed as $\qposList=(\qpos{0},\dots,\qpos{\njoint-1})$. 
	  
	  The pulley radius at each joint $i$ is denoted by $r_i$.
	  Note that for model simplification and to reduce the search space, all tendons passing through the same joint share the same pulley radius.
	  Detailed tendon parameters are represented as $\tendon{0}, \tendon{1},\dots, \tendon{\ntendon-1}$. 
	  When $\tendon{j}$'s start point exists on $\link{i}$, we define $\tendonStart{j}=i$. 
	  Similarly, $\tendonEnd{j}=i$ indicates the end point on $\link{i}$. 
	  These indices must satisfy $0\leq\tendonStart{j}<\tendonEnd{j}\leq \njoint$, meaning start points are positioned proximally and within the arm's range. 
	  
\begin{figure}[t]
		  \centering
		  \includegraphics{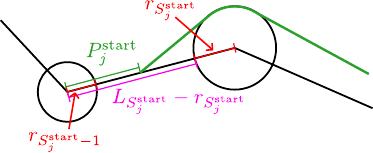}
		  \caption{The range of $\tendonStartX{j}$ is limited to $r_{\tendonStart{j}-1} < \tendonStartX{j} < L_{\tendonStart{j}}-r_{\tendonStart{j}}$ by the radii of the pulleys at both ends of the $\link{\tendonStart{j}}$.}
		  \label{fig:tendon_range}
	  \end{figure}

	  The start point position is expressed as $\tendonStartX{j}\in\RealNumber$, 
	  representing the distance from the base of $\link{\tendonStart{j}}$ to $\tendon{j}$'s origin point (where the actuator or global transmission unit is assumed to be anchored).
	  To avoid pulley interference, when $\tendonStart{j}>0$, $\tendonStartX{j} > r_{\tendonStart{j}-1}$, 
	  and when $\tendonEnd{j}<\njoint$, $\tendonStartX{j} < L_{\tendonStart{j}} - r_{\tendonStart{j}}$. 
	  This configuration is depicted in \figref{fig:tendon_range}. 
	  Similarly, the end point position $\tendonEndX{j}$ follows analogous constraints. 
	  Finally, we define tendon-joint positional relations through $\tendonSide{i}{j}$: 
	  1 indicates $\tendon{j}$ passes above joint $i$, while 0 denotes passing below. 
	  In \figref{fig:model}'s example, $\tendonSide{0}{j}=\tendonSide{1}{j}=1$. 
	  
	  Thus, the arm model is fully defined by specifying:
	  \begin{itemize}
		  \item Link lengths $L_i$, joint angles $\qposList$, pulley radii $r_i$,
		  \item Start point link indices $\tendonStart{j}$, start point positions $\tendonStartX{j}$,
		    end point link indices $\tendonEnd{j}$, and end point positions $\tendonEndX{j}$ for each tendon, 
		  \item Wrapping relationships $\tendonSide{i}{j}$ between tendons and pulleys.
	  \end{itemize}
	  \begin{figure}[t]
		  \centering
		  \includegraphics{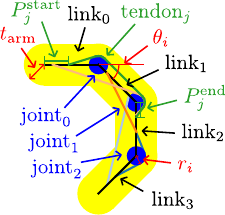}
		  \caption{The example of the model of the arm.
			  The number of joints is $\njoint=3$ and the number of tendons is $\ntendon=5$.}
		  \label{fig:model}
	  \end{figure}
  
  \subsection{Calculation of muscle Jacobian considering tendon shortcut}
	  \label{subsec:jacobian}
	  \newcommand{\isWrap}[1]{\text{is\_wrap}_{#1}}
	  In this section, we describe the method for calculating muscle Jacobians considering tendon shortcuts. 
	  The muscle Jacobian $G(\qposList) \in \RealNumber^{\njoint \times \ntendon}$ is defined component-wise as follows:
	  \begin{equation}
		  G(\qposList)_{i,j} = \frac{\partial l_j}{\partial \qpos{i}}
	  \end{equation}
	  where $l_j$ represents the length of tendon $\tendon{j}$. 
	  To compute this, it is necessary to calculate the specific path taken by each $\tendon{j}$ under given $\qposList$, 
	  particularly considering their wrapping relationships $\tendonSide{i}{j}$. 
	  
\begin{figure}[t]
		  \centering
		  \includegraphics{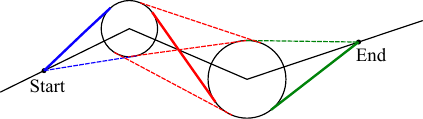}
		  \caption{The example of the tangent line between pulleys.
			  This is the case when the tendon passes through the upper side of the left pulley
			  and the lower side of the right pulley.
			  The red line represents the tendon between the two pulleys, 
			  the blue line represents the tendon between the starting point and the pulley, 
			  and the green line represents the tendon between the end point and the pulley.
			  The dashed line represents a common tangent line other than the one the tendon actually passes through.
		  }
		  \label{fig:tangent_line}
	  \end{figure}

	  First, we determine the wrapping state of each pulley.
	  For each joint $i$ in the tendon's range ($\tendonStart{j} \leq i < \tendonEnd{j}$), we introduce a binary variable $\isWrap{i}$.
	  We set $\isWrap{i}=1$ if the tendon is assumed to wrap around the pulley at joint $i$, and $\isWrap{i}=0$ if it shortcuts the pulley.
	  This determination is made for all indices satisfying $\tendonStart{j} \leq i < \tendonEnd{j}$. 
	  Under these assumptions, the tendon's path is computed. 
	  Note that in this study, we assume that the tendon does not wrap around any pulley for more than a full revolution ($2\pi$ rad).
	  Pulleys assumed to not wrap (i.e., $\isWrap{i}=0$) have no effect on the tendon and can be disregarded. 
	  For pulleys where wrapping is assumed ($\isWrap{i}=1$), we calculate the tendon's wrapped path.  
	  The computation of the tendon's path reduces to finding common tangents between circles. 
	  Between two adjacent pulleys, the tendon's path corresponds to one of the common external or internal tangents between their circular radii. 
	  For paths involving start/end points and pulleys, we treat them as tangents from a point to a circle. 
	  By assuming zero-radius pulleys at the start/end points, these cases also reduce to finding common tangents between circles.  
	  The specific choice among the four possible common tangents between pulleys is determined by the combination of $\tendonSide{i}{j}$ values at each joint's pulley. 
	  As illustrated in \figref{fig:tangent_line}, examples of tendon paths between pulleys can be derived this way. 
	This process allows us to compute the tendon's path segment between any two adjacent assumed-wrapped pulleys along its path.
	  
\begin{figure}[t]
		  \centering
		  \subfloat[These are examples of checks to see if the assumption that the tendon wraps around the pulley is correct.
			  This is the case when the tendon passes under the center pulley.
			  In the example on the left, the assumption that the tendon wraps around is correct because the line segments on both sides do not intersect.
			  In the example on the right, the assumption that the tendon wraps around is incorrect because the two line segments intersect each other, 
			  and the assumption that the tendon does not actually wrap around is correct, as shown by the dashed line.]{%
			  \resizebox*{6.75cm}{!}{\includegraphics{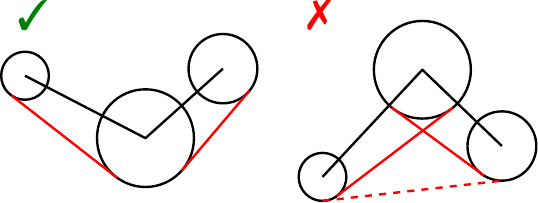}
				  \label{fig:tangent_pulley_check}
			  }
		  }\hspace{5pt}
		  \subfloat[These are examples of checks to see if the assumption that the tendon does not wrap around the pulley is correct.
			  This is the case when the tendon passes over the pulley.
			  In the example on the left, the assumption that the tendon does not wrap around the pulley is correct because the tendon does not intersect the pulley but passes over the top of the joint.
			  In the middle example, the tendon has intersected the pulley, 
			  and in the right example, the assumption that the tendon does not wrap around is incorrect because the tendon passes below the joint. 
			  The assumption that the tendon wraps around is correct, as shown by the dashed line.]{%
			  \resizebox*{6.75cm}{!}{\includegraphics{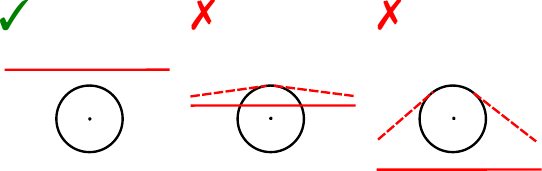}
				  \label{fig:intangent_pulley_check}
			  }
		  }

		  \caption{The examples on checking assumptions about whether the tendon wraps around each pulley.}
	  \end{figure}

	  Next, we verify the validity of the wrapping assumptions $\isWrap{i}$ set for the pulleys. 
	  A valid assumption must satisfy two conditions:
	  \begin{itemize}
		  \item The tendon path is the shortest possible route.
		  \item The tendon path maintains positional consistency with $\tendonSide{i}{j}$ relative to each pulley.
	  \end{itemize}
	  For any assumed-wrapped pulley, its two adjacent tangent lines must not intersect (\figref{fig:tangent_pulley_check}). 
	  If intersection occurs, it implies a non-shortest path route invalidating the assumption.  
	  For pulleys assumed unwrapped, the connecting tangent between neighboring assumed-wrapped pulleys must not intersect this pulley,
	and the positional relationship between the pulley center and tangent line must align with $\tendonSide{i}{j}$ (\figref{fig:intangent_pulley_check}).
	  Violation of either condition invalidates the assumption.  
	  This validation process is formalized in Algorithm \ref{alg:wrap_check}. 
	  By systematically checking all $2^{\tendonEnd{j}-\tendonStart{j}}$ possible assumption combinations within the index range $\tendonStart{j}\leq i < \tendonEnd{j}$, 
	  we can uniquely determine the valid wrapping configuration $\isWrap{i}$ and the corresponding tendon path considering both wrapping and shortcut.
	  The exhaustive search guarantees identification of the physically realizable tendon path with correct geometrical constraints.
	  
	  Finally, based on the obtained tendon path, we compute the muscle Jacobian $G(\qposList)$.  
	  For joints unaffected by $\tendon{j}$ (i.e., when $i < \tendonStart{j}$ or $\tendonEnd{j} \leq i$),  
	  the corresponding component of the muscle Jacobian is $G(\qposList)_{i,j} = 0$
	  When $\tendon{j}$ is wrapped around pulley of $\joint{i}$, the component is determined by:  
	  \begin{equation}
		  G(\qposList)_{i,j} = \begin{cases}
			  -r_i & \text{if } \tendonSide{i}{j} = 1  \\
			  r_i  & \text{if } \tendonSide{i}{j} = 0
		  \end{cases}
	  \end{equation}
	  When $\tendon{j}$ is not wrapped around pulley of $\joint{i}$, 
	  we use the distance $d$ between the $\joint{i}$ and the tangent line:  
	  \begin{equation}
		  G(\qposList)_{i,j} = \begin{cases}
			  -d & \text{if } \tendonSide{i}{j} = 1  \\
			  d  & \text{if } \tendonSide{i}{j} = 0
		  \end{cases}
	  \end{equation}
	  This completes the computation of all components $G(\qposList)_{i,j}$ for the shortcut-aware muscle Jacobian matrix, 
	  accounting for both pulley-wrapping and tangent-line geometry in tendon path calculations.
	  
	  \begin{figure}[t]
		  \tikzset{
			  algpxIndentLine/.style={%
					  draw=none,          
					  opacity=0           
				  }
		  }
		  \begin{algorithm}[H]
			  \caption{The algorithm to check if $\isWrap{i}$, the assumption about which pulley the tendon wraps around, is correct.}
			  \label{alg:wrap_check}
			  \begin{algorithmic}[1]
				  \Function{Check}{$\isWrap{i}$}
				  \For {$i = \tendonStart{j}$ to $\tendonEnd{j}-1$}
				  \If {$\isWrap{i} = 1$}
				  \If {two segments on the pulley intersect each other}
				  \LComment {Case of the right of \figref{fig:tangent_pulley_check}}
				  
				  \State \Return False
				  \EndIf							
				  \Else[$\isWrap{i} = 0$]
				  \If {the segment intersects the pulley}
				  \LComment {Case of the center of \figref{fig:intangent_pulley_check}}
				  \State \Return False
				  \EndIf
				  \If {the segment is on the wrong side of the pulley}
				  \LComment {Case of the right of \figref{fig:intangent_pulley_check}}
				  \State \Return False
				  \EndIf
				  \EndIf
				  \EndFor
				  \State \Return True
				  \EndFunction
			  \end{algorithmic}
		  \end{algorithm}
	  \end{figure}
  \subsection{Evaluation of joint torque}
	  \label{subsec:torque}
	  \newcommand{\fmin}[1]{F_{\text{min},#1}}
	  \newcommand{\fmax}[1]{F_{\text{max},#1}}
	  The method for evaluating the joint torques of an arm system is described here. 
	  In tendon-driven arms, the joint torque can be expressed by the following equation:  
	  \begin{equation}  
		  \bm{\tau} = -G(\qposList)^T \bm{f}
	  \end{equation}
	  Here, $\bm{\tau}\in\RealNumber^{\njoint}$ denotes the joint torque vector, 
	  $G(\qposList)$ is the arm's muscle Jacobian derived in \secref{subsec:jacobian}, 
	  and $\bm{f}\in\RealNumber^{\ntendon}$ represents the tendon tension vector.  
	  Let the range of exertable tension for each tendon be denoted as $[\fmin{j}, \fmax{j}]$. 
	  The set of all possible tendon tensions is then defined as:  
	  $\mathbb{F} = \prod_{j=1}^{\ntendon}[\fmin{j}, \fmax{j}] = \{\bm{f} \in \RealNumber^{\ntendon} | \fmin{j} \leq \bm{f}_j \leq \fmax{j} \}$.
	  Therefore, the set of possible joint torques is:  
	  $\mathbb{T} = \{\bm{\tau} = -G(\qposList)^T \bm{f} | \bm{f} \in \mathbb{F} \}$.
	  
	  The ability of an arm to exert torques in all directions is equivalent to its joint torque workspace $\mathbb{T}$ encompassing a hypersphere centered at the origin. 
	  A larger contained hypersphere indicates greater capability for the arm to freely achieve joint torques across orientations. 
	  The radius of the largest hypersphere centered at the origin that $\mathbb{T}$ contains can be determined by finding the shortest distance $d_S$ between the surface $S$ of $\mathbb{T}$ and the origin. 
	  Conversely, if the origin lies outside $\mathbb{T}$, this implies the existence of joint torque directions the arm cannot realize in certain configurations. 
	  In such cases, we treat $d_S$ as a negative value.
	  
	  We describe the method for determining the shortest distance between $S$ and the origin.  
	  First, $\mathbb{F}$ can be considered as the convex hull of the vertex set $\mathbb{F}_p = \prod_{j=1}^{\ntendon} \{ \fmin{j}, \fmax{j} \}$. 
	  Since $\mathbb{T}$ is obtained through a linear transformation of $\mathbb{F}$, it corresponds to the convex hull of the transformed vertex set $\mathbb{T}_p = \{ -G(\qposList)^T \bm{f} | \bm{f} \in \mathbb{F}_p \}$. 
	  By evaluating all $2^{\ntendon}$ vertices formed by selecting either the minimum or maximum tendon tension for each of the $\ntendon$ tendons, 
	  applying the transformation $-G(\qposList)^T$ to these vertex points yields $\mathbb{T}_p$. 
	  The convex hull constructed from these transformed vertices is denoted as $C$, which represents the feasible joint torque space $\mathbb{T}$. 
	  Its bounding faces constitute the surface $S$, which can be used to compute the shortest distance to the origin.
	  
	  $C$ can be represented using the equation of the constituting surface. 
	  If the surface equation is given by $\bm{a}^T\bm{x} + b = 0$, 
	  then by appropriately selecting the coefficients, 
	  we can obtain a signed distance $\bm{a}^T\bm{x} + b$ where the interior side of $C$ is positive. 
	  For points that have non-negative distances to all surfaces, those points are contained within $C$. 
	  For points contained in $C$, the minimum value of the distances between the point and each surface corresponds to the distance between the point and $S$. 
	  For points not contained in $C$, the negative of the minimum distance between the point and the points contained in $C$ corresponds to the distance between the point and $S$. 
	  By computing the distance from the origin according to the above, we can obtain $d_S$.
	  
	  By computing $d_S$ for all possible $\qposList$ and evaluating the minimum value, 
	  we can assess the joint torque of the arm. 
	  However, in this study, to reduce computational costs, 
	  we compute $d_S$ for a representative set of multiple $\qposList$ and evaluate the joint torque based on the minimum value among them.
  \subsection{Evaluation of Arm Thickness}
	  \newcommand{\thickness}{t_{\text{arm}}}
	  In this section, we describe the method for evaluating the thickness of the arm. 
	  Let $\bm{p} \in \RealNumber^2$ be an arbitrary point in the workspace.
	  We define $d_{arm}(\bm{p})$ as the shortest Euclidean distance between the arm's link structure (the set of line segments representing links) and the point $\bm{p}$. 
	  Note that each $\link{i}$ of the arm is considered as a line segment of length $L_i$, 
	  and the minimum value of the shortest distances between each line segment and $\bm{p}$ is defined as $d_{arm}(\bm{p})$. 
	  We evaluate the arm thickness by considering the distance of points located on the surface of pulleys and tendons from the links.
	  Specifically, for the set of points $\mathcal{P}_{components}$ constituting the pulleys and tendons,
	  the maximum value of these distances is defined as the thickness of the arm, denoted by $\thickness = \max_{\bm{p} \in \mathcal{P}_{components}} d_{arm}(\bm{p})$.
	  
	  For pulleys, the radius of the pulley corresponds to the distance between the pulley and the arm. 
	  For tendons, the portion wrapped around the pulley has a distance equal to the pulley's radius, 
	  while the unwrapped portion is considered as a line segment. 
	  We sample multiple points on this line segment and compute the distance to the arm at each point, taking the maximum value. 
	  By computing this distance for all line segments constituting the tendon and taking the maximum value, 
	  we obtain the maximum distance between the points constituting the tendon and the arm. 
	  Finally, by computing the maximum value of the distances between each pulley and each tendon and the arm, 
	  we can determine the thickness of the arm, $\thickness$. 
	  
	  By computing $\thickness$ for all possible $\qposList$ and evaluating the minimum value, we can assess the thickness of the arm. 
	  However, in this study, to reduce computational costs, 
	  we compute $\thickness$ for a representative set of multiple $\qposList$ and evaluate the thickness based on the minimum value among them.
  
  \subsection{Optimization of Tendon Position}
	  We describe the method for optimizing the tendon configuration. 
	  The number of arm joints $\njoint$ and tendons $\ntendon$, as well as the link length $L_i$, are fixed. 
	  The parameters to be optimized are $r_i$, $\tendonStart{j}$, $\tendonEnd{j}$, $\tendonStartX{j}$, $\tendonEndX{j}$, and $\tendonSide{i}{j}$. 
	  The objective functions are two-fold: $d_S$ and $\thickness$. 
	  We aim to maximize $d_S$ and minimize $\thickness$. 
	  
	  Additionally, all variables representing length and force are computed as dimensionless quantities. 
	  Since $d_S$ is proportional to the reference length and force, and $\thickness$ is proportional to the reference length, 
	  multiplying the reference length and force by a constant factor does not affect the optimization result. 
	  By computing the variables as dimensionless quantities, 
	  the optimization results can be applied to arms of various scales with different link lengths and exertable forces.
\section{EXPERIMENTS}
  \label{sec:experiments}
  \subsection{Experimental Setting}
	  \label{subsec:experimental_setting}
\begin{figure}[t]
		  \centering
		  \includegraphics{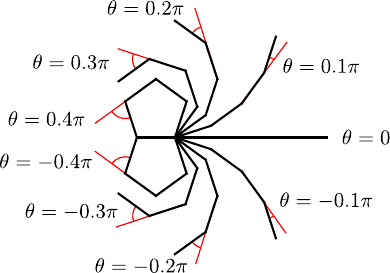}
		  \caption{The example of theta listing for a 4-joint case.
			  The angles shown in the figure go into all elements of theta.}
		  \label{fig:theta_list}
	  \end{figure}

	  We describe the experimental setup of this study. 
	  The length of each link in the arm was set to $L_i=1$. 
	  The pulley radius was limited to $0 \leq r_i \leq 0.3$ to prevent mutual interference. 
	  The range of exertable tension for each tendon was set to $[\fmin{j}, \fmax{j}] = [0, 1]$. 
	  The evaluation of $d_S$ and $\thickness$ was performed by substituting $\theta$ values sampled at equal intervals from $[-\frac{2\pi}{\njoint+1},\frac{2\pi}{\njoint+1}]$ into each $\qpos{i}$, resulting in 9 different $\qposList$. 
	  An example for 4 joints is shown in \figref{fig:theta_list}. 
	  
	  For multi-objective optimization, we employed the NSGA-II (Non-dominated Sorting Genetic Algorithm II) \cite{Deb2002NSGAII}, a widely used evolutionary algorithm capable of efficiently exploring Pareto-optimal solutions.
	  We used the implementation provided by the Optuna library \cite{Optuna}. 
	  The arm model calculation utilized the numerical computation library JAX\cite{JAX}. 
	  For the convex hull calculation in joint torque evaluation, we employed the ConvexHull class included in SciPy\cite{SciPy}.
	  
	  \begin{figure}[t]
		  \centering
		  \subfloat[The optimization result of the $\njoint=2$ and $\ntendon=3$ case.]{%
			  \resizebox*{6cm}{!}{\includegraphics{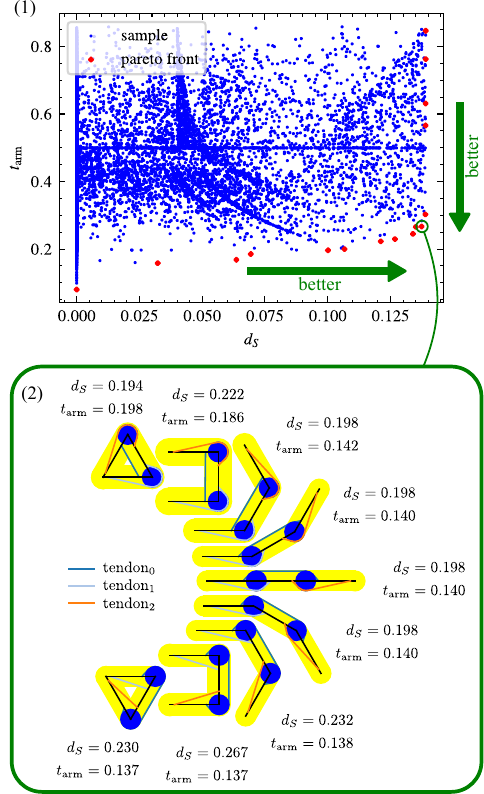}}
			  \label{fig:2joints_result}
		  }\hspace{5pt}
		  \subfloat[The optimization result of the $\njoint=4$ and $\ntendon=8$ case.]{%
			  \resizebox*{6cm}{!}{\includegraphics{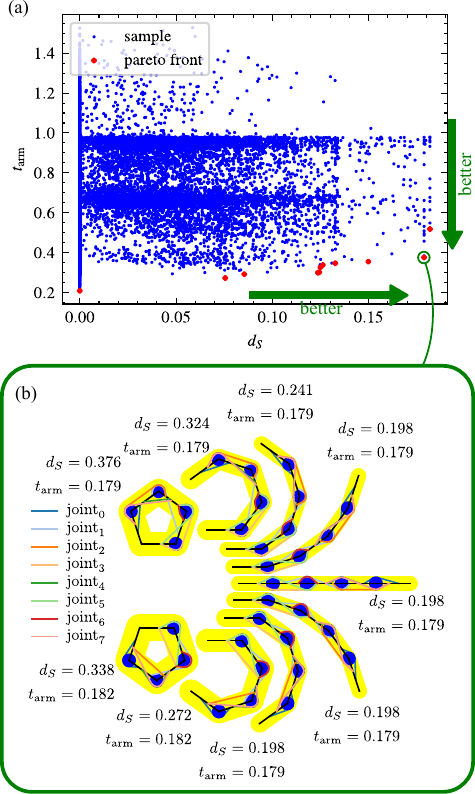}}
			  \label{fig:4joints_result}
		  }
		  \caption{(1) The result of the multi-objective optimization.
			  The blue points are the sampling points, and the red points are the Pareto fronts.
			  The horizontal axis is the minimum distance between the surface of the demonstrable joint torque space and the origin in each posture, 
			  and the vertical axis is the arm thickness.
			  (2) An example of a solution with both $d_S$ and $\thickness$ moderately superior among Pareto fronts.
			  It shows the arm's posture and tendons at each $\qposList$, its posture-specific $d_S$, and $\thickness$.
			  The blue circles indicate pulleys and the yellow areas indicate areas where the distance from the arm is less than the arm's thickness.
		  }
		  \label{fig:optimization_results_show}
	  \end{figure}
	  \begin{figure}[t]
		  \centering
		  \subfloat[Two joints arm with 3--6 tendons.]{%
			  \resizebox*{6.75cm}{!}{\includegraphics{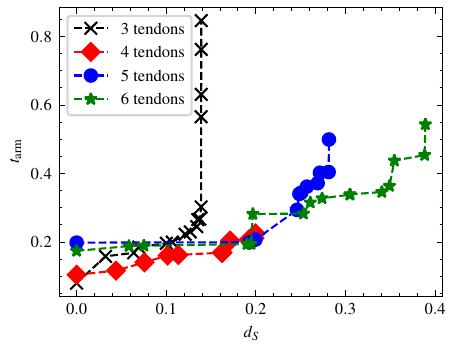}
				  \label{fig:2joints_pareto}
			  }
		  }\hspace{5pt}
		  \subfloat[Three joints arm with 4--7 tendons.]{%
			  \resizebox*{6.75cm}{!}{\includegraphics{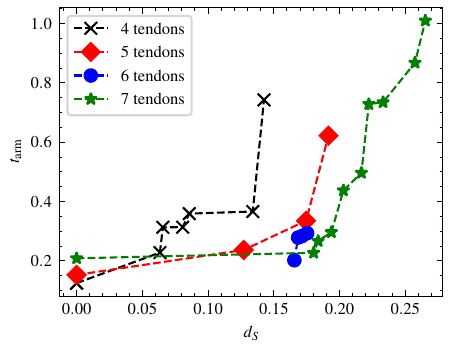}
				  \label{fig:3joints_pareto}
			  }
		  } \\
		  \subfloat[Four joints arm with 5--8 of tendons]{%
			  \resizebox*{6.75cm}{!}{\includegraphics{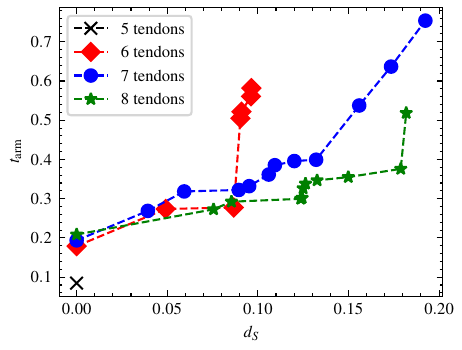}
				  \label{fig:4joints_pareto}
			  }
		  }\hspace{5pt}
			\subfloat[Five joints arm with 6--9 tendons. ``9 tendons+'' represents a case with 40,000 samples.]{%
			  \resizebox*{6.75cm}{!}{\includegraphics{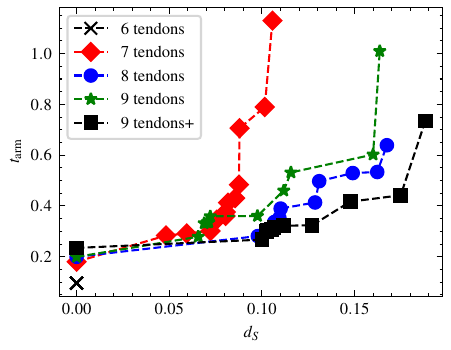}
				  \label{fig:5joints_pareto}
			  }
		  }
		  \caption{
			  The optimization result of the multi joints arm with different number of tendons.
			  Only Pareto solutions are plotted.
		  }
		  \label{fig:multi_pareto}
	  \end{figure}
  \subsection{Easy Configuration: Two Joints Arm with Three Tendons}
	  \label{subsec:easy_configuration}
	  We describe the optimization results for a 2-joint arm.
	  In this experiment, we set the number of joints to $\njoint=2$ and the number of tendons to $\ntendon=3$, and performed optimization with a sample size of 20000. 
	  The optimization results are shown in \figref{fig:2joints_result}. 
	  
	  In \figref{fig:2joints_result}(1), we plot the sampled points and Pareto solutions during multi-objective optimization, 
	  with $d_S$ on the x-axis and $\thickness$ on the y-axis. 
	  Therefore, better solutions are located at the lower right of the figure. 
	  From this, it can be seen that several solutions are obtained as Pareto solutions, 
	  ranging from those with larger $d_S$ to those with smaller $\thickness$. 
	  However, the solution with the smallest $\thickness$ has $d_S=0$, which is a meaningless design solution. 
	  Additionally, among the solutions with large $d_S$, there are some where $\thickness$ suddenly increases rapidly as $d_S$ increases. 
	  
	Therefore, we consider solutions with large $d_S$, low $\thickness$, and a knee point near the trade-off corner to be desirable.
	To make this idea concrete, we select one representative solution using the following rule in this study: the Pareto solution that lies farthest from the straight line connecting the two extreme Pareto solutions in the $(d_S,\thickness)$ plane.
	These two extreme solutions correspond to the trade-off endpoints in this study: one emphasizes larger $d_S$, and the other emphasizes smaller $\thickness$.
	This criterion is used only as a representative example in this study, and it does not mean that this particular solution must always be selected.
	  Thus, we show an example of such a Pareto solution in \figref{fig:2joints_result}(2). 
	  In this example, $\tendon{1}$ and $\tendon{2}$ wrap around $\joint{0}$ and $\joint{1}$ from the bottom, 
	  while $\tendon{0}$ wraps around both joints from the top. 
	  Specifically, $\tendonSide{i}{j}$ is as follows:
	  \begin{equation}
		  \text{tendon\_side} = \left(
		  \begin{array}{ccc}
				  1 & 0 & -  \\
				  1 & - & 0
			  \end{array}
		  \right)
	  \end{equation}
	  Here, the "-" symbol indicates that the tendon does not cross the joint. 
	  This configuration is also easily conceived by humans, 
	  but it can be seen that optimization can also generate such configurations. 
	  Furthermore, when the joint angle becomes large, some tendons are slightly shortcut from the pulley and wrapping.
	  It can be seen that they are increasing the joint torque while avoiding excessive thickness. 
	  This is an important consideration in actual arm design, taking into account the wrapping and shortcut of tendons.
  \subsection{Complex Configuration: Four Joints Arm with Eight Tendons}
	  \label{subsec:complex_configuration}
	  Similar to \secref{subsec:easy_configuration}, 
	  we conducted optimization under more complex conditions, i.e., for a 4-joint arm. 
	  The number of joints in the arm was set to $\njoint=4$ and the number of tendons to $\ntendon=8$. 
	  In the optimization process, the sample number was set to 20000. 
	  An example of the obtained optimal solution and Pareto solution is shown in \figref{fig:4joints_result}. 
	  
	  The $\tendonSide{i}{j}$ in the example is represented in the following matrix form:
	  \begin{equation}
		  \text{tendon\_side} = \left(
		  \begin{array}{cccccccc}
				  - & - & - & 0 & - & 1 & - & -  \\
				  - & 1 & - & 0 & - & 0 & 1 & -  \\
				  - & 0 & 0 & - & 1 & - & - & 1  \\
				  1 & - & 0 & - & - & - & - & 0
			  \end{array}
		  \right)
	  \end{equation}
	  
	  We will provide a detailed discussion of this configuration example. 
	  From the matrix composition, it can be seen that tendons spanning one joint and those spanning two joints coexist. 
	  As for the tendons spanning only one joint, $\tendon{0}, \tendon{4},$ and $\tendon{6}$ exist, all of which pass through the upper side. 
	  These configurations contribute to the exertion of torque by the arm. 
	  
	  Among the tendons spanning two joints, $\tendon{2}$ and $\tendon{3}$ pass through the same side (lower side) at two joints 
	  and cause a shortcut when the joint angle becomes large in the negative direction. 
	  In combination with the fact that the tendons spanning one joint pass through the upper side, 
	  these configurations enable efficient exertion of joint torque in both directions. 
	  
	  Furthermore, $\tendon{1}, \tendon{5},$ and $\tendon{7}$ pass through different sides at two joints and evenly span each pair of joints, 
	  i.e., $\joint{0}$ and $\joint{1}$, $\joint{1}$ and $\joint{2}$, and $\joint{2}$ and $\joint{3}$. 
	  Since these configurations suppress the shortcut of the arm's links, 
	  it is possible to effectively increase the torque at each joint without increasing the thickness. 
	  
	  Additionally, the fact that there are no tendons spanning three or more joints is considered to be an optimization result for suppressing the arm's thickness. 
	  Since it is a 4-joint arm and the number of tendons is as many as 8, proper design by humans would be difficult. 
	  However, through multi-objective optimization, it can be confirmed that a proper configuration considering both joint torque and arm thickness is generated. 
	  Moreover, the design becomes a complex configuration that people would not easily think of, 
	  suggesting the potential ability of optimization methods to explore beyond human intuition. 
  \subsection{Comparison on Multi Configurations}
	  \label{subsec:multi_configuration}
	  Similar to \secref{subsec:easy_configuration} and \secref{subsec:complex_configuration}, 
	  we conducted multiple optimizations while changing the number of joints $\njoint$ and tendons $\ntendon$. 
	  In all cases, the sample number for optimization was set to 20000. 
	  The results are shown in \figref{fig:multi_pareto}. 
	  Since there are $\njoint$ degrees of freedom at the joints, at least $\njoint+1$ tendons are necessary. 
	  Taking this into account, we varied $\ntendon$ from $\njoint+1$ to $\njoint+4$ and performed optimizations. 
	  
	  We will discuss these results. 
	  First, it can be seen that for most $\njoint$ and $\ntendon$, 
	  Pareto solutions considering the trade-off between $d_S$ and $\thickness$ have been obtained. 
	  However, for $(\njoint, \ntendon) = (4,5)$ and $(5,6)$, only solutions with $d_S=0$ were obtained. 
	  This is likely due to the fact that there are very few designs in the search space that can exert torque. 
	  
	  Additionally, for $\njoint=2,3,4$, as $\ntendon$ increases, $d_S$ tends to increase for the same $\thickness$. 
	  This result makes sense, since increasing the number of tendons increases the total force that can be exerted. 
	  On the other hand, for $\njoint=5$, $d_S$ shows a valid trade-off curve. With the standard sample size of 20,000, the improvement trend at $\ntendon=9$ is less pronounced compared to lower dimensions, suggesting the optimization has approached its exploration limits at this dimensionality.
	  This result was intentionally included to demonstrate the upper limit of the design space that can be effectively explored with a fixed sample size of 20,000.
	  However, when we increased the sample size to 40,000 for the 5-joint, 9-tendon case, the performance of 9 tendons exceeded that of 8 tendons, demonstrating that the 20,000-sample optimization was indeed insufficient for this high-dimensional problem. 
	  This improvement confirms that for the 5-joint system, increasing the number of tendons does enhance the achievable torque capability, consistent with the trend observed in lower-dimensional cases, provided sufficient optimization samples are available.
	  The curse of dimensionality in high-DOF systems remains evident: simply increasing samples is inefficient at such complexity levels, pointing towards the need for more scalable optimization strategies in future work.
	  
	  Based on the above, it has been shown that through multi-objective optimization, 
	  optimal tendon configurations considering the trade-off between joint torque and arm thickness can be obtained for various numbers of joints and tendons. 
	  These results provide a comprehensive view of the design landscape across different complexities, including the boundaries of exploration feasibility. 
\section{DISCUSSION}
  \label{sec:discussion}
  
  The optimization results demonstrate that the proposed method effectively navigates the trade-off between maximizing joint torque and minimizing arm thickness. 
  The obtained Pareto fronts provide a clear visualization of potential design candidates, ranging from compact arms with moderate torque to high-torque designs with necessary thickness.
  
  In the designs obtained from \secref{subsec:easy_configuration} and \secref{subsec:complex_configuration}, we observed a mix of tendons spanning single joints and multiple joints.
  Tendons spanning two joints tended to shortcut across the joint when the angle increased, significantly boosting the moment arm and thus the joint torque.
  Conversely, single-joint tendons contributed to independent torque control, enhancing the versatility of the torque workspace.
  This observation aligns with geometric intuition but provides specific, optimized parameters (start/end positions) that are difficult to determine manually.
  Compared to the heuristic designs often used, our method optimizes the exact attachment points, which directly influence the moment arm profiles.
  For instance, the optimal start points were found to be located such that the moment arm is maximized at the specific joint angles where the torque requirement is highest or the manipulability needs to be maintained.
  This level of detail in design parameters goes beyond simple routing type selection.
  
  Regarding the choice of solutions from the Pareto set, designers can select an optimal configuration based on specific constraints. 
  For example, if an application requires the arm to enter a $50$mm gap, a solution can be selected from the region where $\thickness < 50$mm, maximizing torque under this constraint. 
  The "knee points" on the Pareto front often represent designs with the best balance, offering significant torque improvements for minimal thickness penalties.
	Comparing our approach to Tsai's isotropic design\cite{Tsai1996Isotropic}, which focuses on uniform force transmission with conventional routing assumptions, our method provides a complementary perspective by explicitly optimizing shortcut-aware tendon routing and the resulting trade-off between compactness and maximum torque capability.
	In this sense, the main contribution of this study is to provide a set of multiple practical design candidates on the Pareto front, from which designers can select according to task-specific constraints.
  
  From the results in \secref{subsec:multi_configuration}, increasing the number of tendons was shown to improve torque performance without necessarily increasing thickness.
  This implies that adding redundancy is an effective strategy for high-performance compact arms.
  
  Finally, regarding the required sample size, as the number of joints and tendons increases, the search space expands exponentially (the curse of dimensionality).
  We intentionally maintained a consistent sample size (20,000) across all experiments to identify the scalability limits of the standard optimization approach.
  The results clearly show that while 20,000 samples are sufficient for lower-dimensional cases, they begin to saturate for higher-dimensional problems (e.g., $\njoint=5$).
  This finding is valuable as it provides a benchmark for future research into more sample-efficient algorithms or heuristic-guided methods necessary for complex musculoskeletal designs.
  
  The current study assumes ideal tendon routing without specific hardware constraints.
  In practice, the placement of actuators or pulleys might be restricted.
  However, the "optimal" locations found here serve as a target for mechanical design, guiding where components should ideally be placed.
 
\section{CONCLUSION}
  \label{sec:conclusion}
  In this study, we proposed a design optimization method for tendon-driven arms, 
  addressing the inherent trade-off between maximizing joint torque and minimizing arm thickness. 
  Our approach explicitly considers the effects of tendon wrapping and shortcutting on the moment arm, aspects often simplified in previous geometric optimizations.
  Through multi-objective optimization, we derived Pareto-optimal solutions that enable designers to select configurations with the minimum necessary thickness for a required torque output, or maximize torque for a given size constraint.
  The results confirmed that under simple conditions, the optimized designs align with established engineering intuition, while for more complex setups, the method discovers non-trivial configurations that balance these contradictory objectives.
  Furthermore, we discussed the challenges associated with high-dimensional search spaces, highlighting the computational demands required to fully explore complex multi-joint systems.
  Future research will focus on incorporating more realistic hardware constraints, such as actuator placement limitations, and exploring sample-efficient algorithms to handle increasingly complex robotic systems.

  \bibliographystyle{include/tfnlm}
  \bibliography{include/main}

\begin{thebibliography}{10}
\providecommand{\url}[1]{\normalfont{#1}}
\providecommand{\urlprefix}{Available from: }

\bibitem{ECCE1}
Marques~HG, Jäntsch~M, Wittmeier~S, et~al. Ecce1: The first of a series of
  anthropomimetic musculoskeletal upper torsos. In: 2010 10th IEEE-RAS
  International Conference on Humanoid Robots; 2010. p. 391--396.

\bibitem{Anthrob}
Jäntsch~M, Wittmeier~S, Dalamagkidis~K, et~al. Anthrob - a printed
  anthropomimetic robot. In: 2013 13th IEEE-RAS International Conference on
  Humanoid Robots (Humanoids); 2013. p. 342--347.

\bibitem{Kengoro}
Asano~Y, Kozuki~T, Ookubo~S, et~al. Human mimetic musculoskeletal humanoid
  kengoro toward real world physically interactive actions. In: 2016 IEEE-RAS
  16th International Conference on Humanoid Robots (Humanoids); 2016. p.
  876--883.

\bibitem{Musashi}
Kawaharazuka~K, Makino~S, Tsuzuki~K, et~al. Component modularized design of
  musculoskeletal humanoid platform musashi to investigate learning control
  systems. In: 2019 IEEE/RSJ International Conference on Intelligent Robots and
  Systems (IROS); 2019. p. 7300--7307.

\bibitem{SuperDragon}
Endo~G, Horigome~A, Takata~A. Super dragon: A 10-m-long-coupled tendon-driven
  articulated manipulator. IEEE Robotics and Automation Letters.
  2019;\hspace{0pt}4(2):934--941.

\bibitem{SAQIEL}
Suzuki~T, Bando~M, Kawaharazuka~K, et~al. Saqiel: Ultra-light and safe
  manipulator with passive 3d wire alignment mechanism. IEEE Robotics and
  Automation Letters. 2024;\hspace{0pt}9(4):3720--3727.

\bibitem{TwistSnake}
Tanaka~K, Hamaya~M. Twist snake: Plastic table-top cable-driven robotic arm
  with all motors located at the base link. In: 2023 IEEE International
  Conference on Robotics and Automation (ICRA); 2023.

\bibitem{kobayashi1998tendon}
Kobayashi~H, Hyodo~K, Ogane~D. On tendon-driven robotic mechanisms with
  redundant tendons. The International Journal of Robotics Research.
  1998;\hspace{0pt}17(5):561--571.

\bibitem{KAWAHARAZUKA2022104067}
Kawaharazuka~K, Nishiura~M, Toshimitsu~Y, et~al. Robust continuous motion
  strategy against muscle rupture using online learning of redundant
  intersensory networks for musculoskeletal humanoids. Robotics and Autonomous
  Systems. 2022;\hspace{0pt}152:104067.
  \urlprefix\url{https://www.sciencedirect.com/science/article/pii/S0921889022000331}.

\bibitem{Tsai1996Isotropic}
Ou~YJ, Tsai~LW. Isotropic design of tendon-driven manipulators. Journal of
  Mechanical Design. 1996 09;\hspace{0pt}118(3):360--366.
  \urlprefix\url{https://doi.org/10.1115/1.2826893}.

\bibitem{2012Asaoka}
Asaoka~T, Kawamura~M, Kumakura~S, et~al. Determining the optimal multiarticular
  muscle arrangement of a musculoskeletal robot for a specific motion using
  dynamics simulation. In: 2012 12th IEEE-RAS International Conference on
  Humanoid Robots (Humanoids 2012); 2012. p. 216--221.

\bibitem{2024Kawaharazuka}
Kawaharazuka~K, Yoshimura~S, Suzuki~T, et~al. Design optimization of wire
  arrangement with variable relay points in numerical simulation for
  tendon-driven robots. IEEE Robotics and Automation Letters.
  2024;\hspace{0pt}9(2):1388--1395.

\bibitem{Deb2002NSGAII}
Deb~K, Pratap~A, Agarwal~S, et~al. A fast and elitist multiobjective genetic
  algorithm: Nsga-ii. IEEE Transactions on Evolutionary Computation.
  2002;\hspace{0pt}6(2):182--197.

\bibitem{Optuna}
Akiba~T, Sano~S, Yanase~T, et~al. Optuna: A next-generation hyperparameter
  optimization framework. In: Proceedings of the 25th ACM SIGKDD International
  Conference on Knowledge Discovery \& Data Mining; New York, NY, USA.
  Association for Computing Machinery; 2019. p. 2623--2631; KDD '19.
  \urlprefix\url{https://doi.org/10.1145/3292500.3330701}.

\bibitem{JAX}
Bradbury~J, Frostig~R, Hawkins~P, et~al. {JAX}: composable transformations of
  {P}ython+{N}um{P}y programs ; 2018.
  \urlprefix\url{http://github.com/jax-ml/jax}.

\bibitem{SciPy}
Virtanen~P, Gommers~R, Oliphant~TE, et~al. {{SciPy} 1.0: Fundamental Algorithms
  for Scientific Computing in Python}. Nature Methods.
  2020;\hspace{0pt}17:261--272.

\end{thebibliography}

\end{document}